\documentclass[runningheads]{llncs}
\usepackage{times}  
\usepackage{helvet}  
\usepackage{courier}  
\usepackage{url}  
\usepackage{graphicx}  
\usepackage{amsmath}
\usepackage{float}
\input{epsf}
\begin{document}
\title{Emergence of Addictive Behaviors in Reinforcement Learning Agents 
 }

\titlerunning{Emergence of Addictive Behaviors in Reinforcement Learning}
%
\author{Vahid Behzadan\inst{1} \and 
Roman V. Yampolskiy\inst{2} \and
Arslan Munir\inst{1}} 

\authorrunning{V. Behzadan et al.}
%
\institute{Kansas State University, Manhattan, KS 66506, USA\\
\email{\{behzadan,amunir\}@ksu.edu}\\
\url{http://blogs.k-state.edu/aisecurityresearch/} \and
University of Louisville, Louisville, KY 40292, USA\\
\email{roman.yampolskiy@louisville.edu}\vspace{-2 mm}}
\maketitle              
\begin{abstract}
 This paper presents a novel approach to the technical analysis of wireheading in intelligent agents. Inspired by the natural analogues of wireheading and their prevalent manifestations, we propose the modeling of such phenomenon in Reinforcement Learning (RL) agents as psychological disorders. In a preliminary step towards evaluating this proposal, we study the feasibility and dynamics of emergent addictive policies in Q-learning agents in the tractable environment of the game of Snake. We consider a slightly modified settings for this game, in which the environment provides a ``drug'' seed alongside the original ``healthy'' seed for the consumption of the snake. We adopt and extend an RL-based model of natural addiction to Q-learning agents in this settings, and derive sufficient parametric conditions for the emergence of addictive behaviors in such agents. Furthermore, we evaluate our theoretical analysis with three sets of simulation-based experiments. The results demonstrate the feasibility of addictive wireheading in RL agents, and provide promising venues of further research on the psychopathological modeling of complex AI safety problems. 

\end{abstract}

\section{Introduction}
A necessary requirement for both current and emerging forms of Artificial Intelligence (AI) is the need for robust specification of objectives to the AI agent. Currently, a prominent framework for goal-based control of intelligent agents is Reinforcement Learning (RL) \cite{sutton2018reinforcement}. At its core, the objective of an RL agent is to optimize its actions such that an externally-generated reward signal is maximized. However, RL agents are prone to various types of AI safety problems, among which wireheading is subject to growing interest \cite{yampolskiy2014utility}. This problem is generally defined as the manifestation of behavioral traits that pursue the maximization of rewards in ways that do not align with the long-term objectives of the system \cite{yampolskiy2016taxonomy}. Considering the roots of this paradigm in neuroscientific literature, Yampolskiy \cite{yampolskiy2014utility} argues that wireheading is common in human behaviors, as manifested in traits such as substance addiction \cite{montague2004computational}. This argument is also supplemented with an investigation of wireheading in AI, leading to the conclusion that wireheading in rational self-improving optimizers is a real and open problem. In recent years, various studies have emphasized on the vitality of this problem in the domain of AI safety (e.g., \cite{amodei2016concrete}), and some have proposed solutions for limited instances of wirehead in RL agents (e.g., \cite{everitt2016avoiding}). Yet, the growing complexity of current and emerging application settings for RL gives rise to the need for tractable approaches to the analysis and mitigation of wireheading in such agents.

In response to this growing complexity, a recent paper by the authors \cite{behzadan2018psychopathological} presents an analogy between AI safety problems and psychological disorders, and proposes the adoption of a psychopathological abstraction to capture the problems arising from the deleterious behaviors of AI agents in a tractable framework based on the available tools and models of psychopathology. In particular, \cite{behzadan2018psychopathological} mentions that the RL framework, which itself is inspired by the neuroscientific models of the dopamine system \cite{sutton2018reinforcement}, has been adopted by neuroscientists to develop models of psychological disorders such as schizophrenia and substance addiction \cite{montague2004computational}. Accordingly, the authors propose to exploit this bidirectional relationship to investigate the complex problems of AI safety.

To study the feasibility of the proposals in \cite{behzadan2018psychopathological}, this paper adopts the RL-based model of substance addiction in natural agents \cite{redish2004addiction} to analyze the problem of wireheading in RL agents. To this end, we investigate the emergence of addictive behaviors in a case study of an RL agent training to play the well-known game of Snake \cite{haikio2002nokia} in an environment that provides a ``drug'' fruit in addition to the typical, healthy seed for the snake. By extending the formulation of \cite{redish2004addiction} to Q-learning, we analyze the sufficient conditions for the emergence of addictive behavior, and verify this theoretical analysis via simulation-based experiments. The remainder of this paper provides the required background on RL and RL-based modeling of addiction, details our theoretical analysis, and presents the experimental results. The paper concludes with remarks on future extensions of our findings.

\section{Background}
This section presents an overview of RL and the relevant terminology, as well as a summary of the work by Redish \cite{redish2004addiction} in modeling addiction within the RL framework. Readers interested in further details of either topics may refer to \cite{sutton2018reinforcement} and \cite{montague2004computational}.

\subsection{Reinforcement Learning}
Reinforcement learning is concerned with agents that interact with an environment and exploit their experiences to optimize a decision-making policy. The generic RL problem can be formally modeled as a Markov Decision Process (MDP), described by the tuple $MDP = (S, A, R, P)$, where $S$ is the set of reachable states in the process, $A$ is the set of available actions, $R$ is the mapping of transitions to the immediate reward, and $P$ represents the transition probabilities (i.e., dynamics), which are initially unknown to RL agents. At any given time-step $t$, the MDP is at a state $s_t\in S$. The RL agent's choice of action at time $t$, $a_t \in A$ causes a transition from $s_t$ to a state $s_{t+1}$ according to the transition probability $P_{s_t , s_{t+1}}^{a_t}$. The agent receives a reward $r_{t+1}$ for choosing the action $a_t$ at state $s_t$. Interactions of the agent with MDP are determined by the policy $\pi$. When such interactions are deterministic, the policy $\pi: S\rightarrow A$ is a mapping between the states and their corresponding actions. A stochastic policy $\pi(s)$ represents the probability distribution of implementing any action $a\in A$ at state $s$. The goal of RL is to learn a policy that maximizes the expected discounted return $E[R_t]$, where $R_t = \sum_{k=0}^{\infty}\gamma^k r_{t+k}$; with $r_t$ denoting the instantaneous reward received at time $t$, and $\gamma$ is a discount factor $\gamma\in [0,1]$. The value of a state $s_t$ is defined as the expected discounted return from $s_t$ following a policy $\pi$, that is, $V^{\pi}(s_t) = E[R_t|s_t, \pi]$. The action-value (Q-value) $Q^{\pi}(s_t, a_t) = E[R_t|s_t,a_t, \pi]$ is the value of state $s_t$ after using action $a_t$ and following a policy $\pi$ thereafter.

As a value function-based solution to the RL problem, the Q-learning method estimates the optimal action policies by using the Bellman formulation $Q_{i+1} (s,a) = \mathbf{E}[R + \gamma \max_a Q_i]$ as the iterative update of a value iteration technique. Practical implementation of Q-learning is commonly based on function approximation of the parametrized Q-function $Q(s,a; \theta) \approx Q^\ast (s,a)$. A common technique for approximating the parametrized non-linear Q-function is via neural network models whose weights correspond to the parameter vector $\theta$. Such neural networks, commonly referred to as Q-networks, are trained such that at every iteration $i$, the following loss function is minimized:
\begin{eqnarray}\label{eq:update}
L_i(\theta_i) = \mathbf{E}_{s, a\sim \rho(.)} [(y_i - Q(s,a,;\theta_i))^2]
\end{eqnarray}

where $y_i = \mathbf{E}[R + \gamma \max_{a'}Q(s',a';\theta_{i-1}) | s,a]$, and $\rho(s,a)$ is a probability distribution over states $s$ and actions $a$. 

\subsection{RL Model of Addiction}
One of the earliest computational models of addiction is the seminal work of Redish in \cite{redish2004addiction}. In this paper, Redish assumes the hypothesis that addictive drugs access the same neurophysiological mechanisms as natural learning systems, which can be modeled through the Temporal-Difference RL (TDRL) algorithm \cite{sutton2018reinforcement}. TDRL learns to predict rewards by minimizing a prediction error (i.e., reward-error signal), which, in the natural brain, is believed to be carried by dopamine. Many addictive substances, such as cocaine, increases the dopamine levels. Redish hypothesizes that this noncompensable drug-induced increase of dopamine may lead to incorrect optimizations in TDRL. Considering that the goal of TDRL is to correctly learn the value of each state ($V(s_t)$), TDRL learns the value function by calculating two equations per each action taken by the agent. If the agent leaves state $s_t$ and enters state $s_{t+1}$ and received the reward $r_{t+1}$, then the corresponding reward-error signal, denoted by $\delta$, is given by:
\begin{equation}
\delta(t+1) = \gamma [R(s_{t+1})+ V(s_{t+1})] - V(s_t)
\end{equation} 
Then, $V(s_t)$ is updates as:
\begin{equation}
V(s_t) \leftarrow V(s_t) + \nu\delta,
\end{equation}
where $\nu$ is a learning rate parameter. The TDRL algorithm stops when the value function correctly predicts the rewards. The value function can be seen as a compensation for the reward, as the change in value in taking action $a_t$ leading to the state transition $s_t \rightarrow s_{t+1}$ counter-balances the reward achieved on entering state $s_{t+1}$. This happens when $\delta = 0$. However, cocaine and other addictive drugs produce a transient surge in dopamine, which can be modeled by the assumption that the drug-induced surge in $\delta$ cannot be compensated by changes in the value. In other words, the effect of addictive drugs is to induce a positive reward-error signal regardless of the change in value function, thus making it impossible for the agent to learn a value function that cancels out this positive error. As a result, the agent learns to assign more value to the states leading to the dopamine surge, thus giving rise to the drug-seeking behavior of addicted agents.

\section{Case Study : RL Addiction in Snake}
To investigate the feasibility of addictive wireheading in RL agents, we consider the game of Snake \cite{haikio2002nokia} for formal and experimental analysis. The most basic form of Snake is played by one player who controls the direction of a constantly-moving snake in a grid, with the goal of consuming as many seeds as possible by running into them with the head of the snake. The seeds appear in random positions of the grid, and consumption of each seed increases the length of the snake. Running into the grid walls or the snake itself results in termination, thus maneuvering becomes progressively more difficult as the snake consumes more seeds. 

In this study, the game is modified to include two types of edible items: one is the classical seed that increases the length of snake $L_s$ by 1 unit, and a ``drug'' fruit that increases $L_s$ by $u$ units. The instantaneous reward values in this setting is defined by:
\begin{equation}\label{eq:reward}
r_t = 
\begin{cases}
r_c & \text{if agent consumes a seed, } \\
k.r_c & \text{if agent consumes a drug, } \\
0  & \text{otherwise}
\end{cases}
\end{equation}

The objective of the agent is to maximize the return, defined as $R = \sum_{t=0}^{T}r_t$, where $T$ is the terminal time of an episode. We adopt the formalism of Q-learning as an instance of the TD-learning approach.

The questions that we target in this study are two-fold: first is to analyze whether addictive behaviors may emerge in a Q-learning agent training in this environment, and second is to establish the parametric boundaries of the reward function for such behavior to emerge. The following section presents a formal analysis of these two problems.

\subsection{Analysis}
First, we define addictive behavior as those that demonstrate compulsive pursuit of trajectories that may maximize short-term rewards, but defy the core objective of maximizing the long-term cumulative reward of the agent. 
At a state $s_d$ where the agent can take action $a_m$ to consume a drug (i.e., move into a cell that contains a drug fruit), the $Q$-value is given by:
\begin{equation}
Q(s_d)= k.r_c + \gamma V(s_{d+1}^m),
\end{equation}
where $\gamma \in [0,1]$ is the discount factor and $V(s_{d+1}^m)$ is the value of the resulting state $s_{d+1}^m$. Alternatively, if the agent takes any action $a_g$ other than $a_m$, the $Q$-value is given by:
\begin{equation}
Q(s_d,a_g) = r_c + \gamma V(s_{d+1}^g) 
\end{equation}
The manifestation of addiction can be formulated as:
\begin{eqnarray}
\gamma V(s_{d+1}^m) < \gamma V(s_{d+1}^g) \label{eq:state}\\
Q(s_d, a_m) > Q(s_d, a_g) \label{eq:Qaddict} 
\end{eqnarray}
Eq.~(\ref{eq:state}) can be reformulated as 
\begin{equation}
V(s_{d+1}^m) = V(s_{d+1}^g) / l_{d+1},
\end{equation}
where $l_{d+1} > 1$. From Eq.~(\ref{eq:Qaddict}) we have:
\begin{equation}
k.r_c + \gamma V(s_{d+1}^g)/l_{d+1} > r_c + \gamma V(s_{d+1}^g)  
\end{equation}
which can be rearranged as:
\begin{equation}
\frac{(k-1).r_c}{\gamma (1 - 1/l_{d+1})} > V(s_{d+1}^g)
\end{equation}
To obtain a sufficient upper bound for emergence of addiction, we find the maximum possible value of $V(s_{d+1}^g)$ as follows: in an $n\times n$ grid, the maximum possible score is achieved when all elements of the grid are filled with the length of the agent. Considering the assumption in Eq.~(\ref{eq:state}), an upper bound for the game score (and hence for state value) is $V_{max} = r_c(n^2 - L_0)$, where $L_0$ is the initial length of the snake. Therefore, a sufficient condition on $k$, $r_c$, and $\gamma$ for manifestation of addiction is:
\begin{equation}\label{eq:multiply}
\frac{(k-1)}{\gamma} > n^2 - L_0 .
\end{equation}
Also, for the condition of Eq.~(\ref{eq:state}) to hold, it is necessary for $k$ to be set such that:
\begin{equation}\label{eq:growth}
k.r_c (n^2 - L_0) / u < r_c (N^2 - L_0) / 1  \\
\implies k/u < 1
\end{equation}

\begin{figure}[h]\label{fig:Snake}
	\centering
	\includegraphics[scale=0.30]{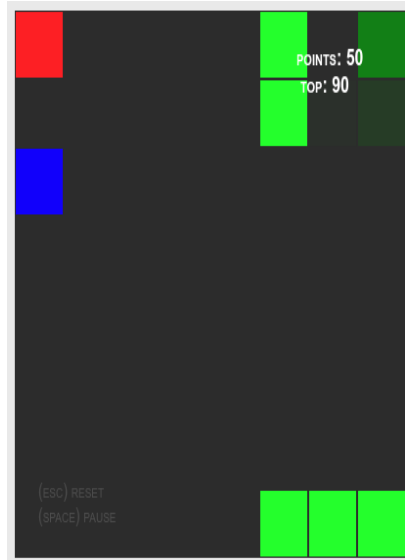}
	\caption{Modified environment of Snake with both healthy seed (red) and drug seed (blue)}
\end{figure}
\section{Experimental Verification}
To evaluate the validity of our analysis, we developed the environment of Snake according to the previously discussed specifications. The environment is comprised of an $n = 8\times 8$ grid, and the initial length of the snake is set to $L_0 = 4$ grid cells. As illustrated in Figure \ref{fig:Snake},  any given time the grid displays two randomly positioned objects, one is the healthy seed (depicted in red), and the other is a drug (colored in blue).

Furthermore, we implemented a tabular Q-learning algorithm with iterative update to train in this environment according to the reward function of Eq.~(\ref{eq:reward}). The exploration mechanism used by our Q-learning implementation is $\epsilon-$greedy, with the initial value of $\epsilon = 0.99$. We consider a constant discount factor $\gamma = 0.9$, and initialize the table of Q-values to $0$. We also consider the instantaneous reward of consuming healthy seeds to be $r_c = 20$. Based on the parametric boundaries derived in Eq.~(\ref{eq:multiply}) and Eq.~(\ref{eq:growth}), we performed three experiments. First, we considered the baseline case where the consumption of drugs do not produce any rewards or length growth (i.e., $k = u = 0$). For the second experiment, we consider a small value of $k = 1.5$, which does not necessarily abide by the sufficient condition of Eq.~(\ref{eq:multiply}). Simultaneously, we set $u = 4$, which does satisfy the condition of Eq.~(\ref{eq:growth}). In the third experiment, we chose $k=6$ and $u=8$ to satisfy both of the derived conditions. To verify the statistical significance of results, the training process of each experiment was repeated 20 times up to $22000$ iterations, and the test-time experiments were repeated 100 times each.

\begin{figure}[h]\label{fig:training}
	\includegraphics[width = \columnwidth]{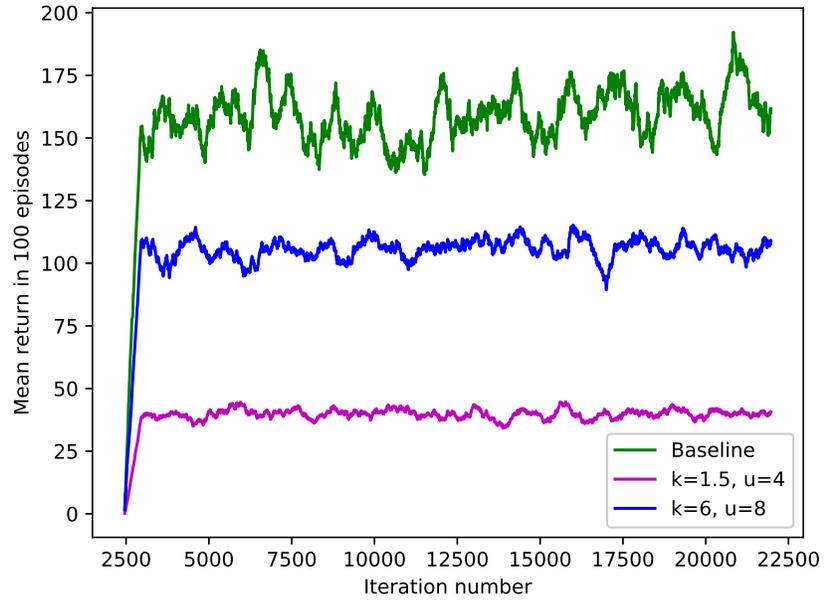}
	\caption{Averaged results of the training process up to 22000 iterations for three experiments}
\end{figure}

\begin{figure}[H]\label{fig:scores}
	\includegraphics[width = \textwidth]{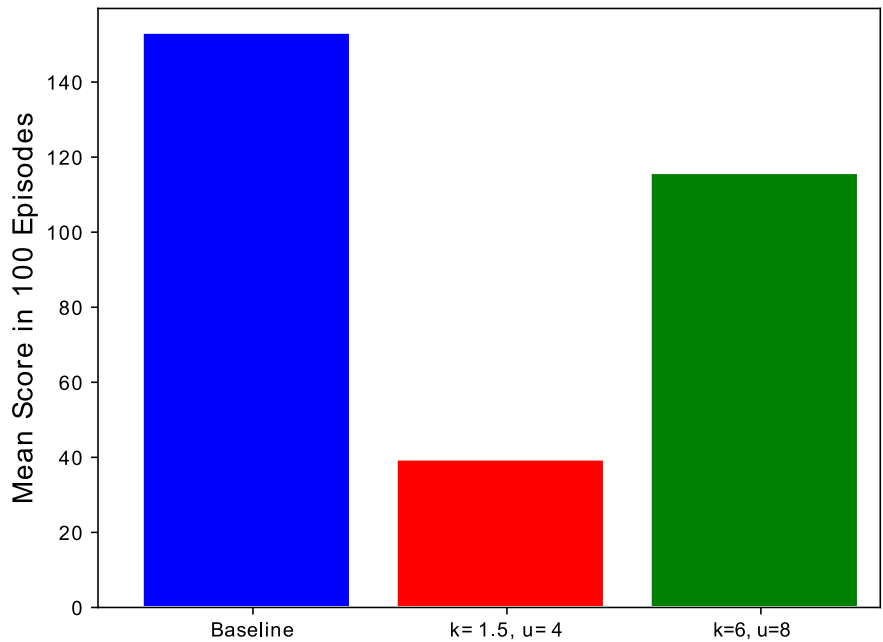}
	\caption{Test-time scores obtained from each experiment}
\end{figure}
\begin{figure}[H]\label{fig:seeds}
	\includegraphics[width = \textwidth]{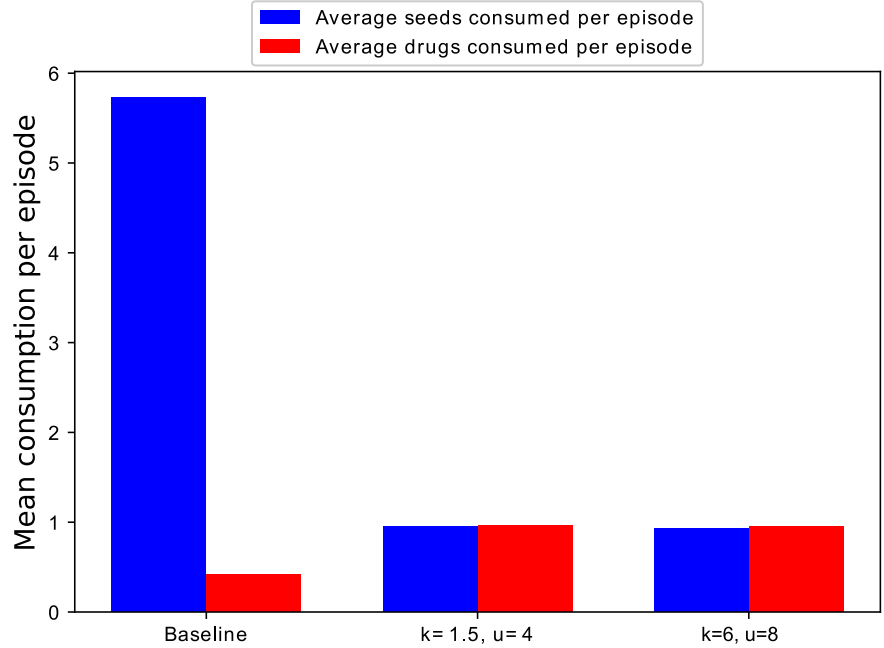}
	\caption{Test-time consumption of healthy seeds and drugs in three experiments}
\end{figure}

Figure \ref{fig:training} demonstrates the training results obtained from the three experiments. It is observed that the baseline case has achieved significantly higher average scores in the same amount of time as the other two cases. Furthermore, the results indicate that the agents training in an environment that includes drug-induced rewards fail to converge towards optimal performance in the observed periods of training. It is also noteworthy that both of the drug-consuming agents reach relatively stable sub-optimal performances in the same period in which the healthy agent enhances its cumulative performance. Moreover, the better performance of the third experiment compared to the second can be explained by the significantly higher instantaneous reward values produced from consuming the drug objects, which noticeably enhance the average performance in comparison to the second experiment with lower values of drug-induced rewards.

The test-time performance of the agents trained in aforementioned environments is illustrated in Figure \ref{fig:scores}. These results are in agreement with those of Figure \ref{fig:training}, as the baseline agents demonstrate superior performance in gaining cumulative rewards, as opposed to the agents trained under drug-induced rewards. Furthermore, Figure \ref{fig:seeds} presents a comparison between the number of healthy seeds and drugs consumed by each agent at test-time. As expected, the baseline results demonstrate a significantly higher consumption of healthy seeds, and the minor levels of drug consumption are due to unintended collisions with the drug objects during game play. It is interesting to note the similarity in the consumption levels of agents trained with drug-induced rewards. In both cases, the agents consume slightly more drugs than healthy seeds, which indicates bias towards the short-term reward-surges of consuming drugs over pursuit of healthy seeds. Although, the difference between the averaged levels of healthy and drug seed consumption is not significant, which may indicate that the agents learned a balanced sub-optimal policy, resulting in confinement within local optima. While this problem can be resolved via enhanced randomization and exploration strategies, one shall consider the effect of this deficiency on sample-efficiency and the consequent limitations of real-world applications.

\section{Conclusion}
We studied the feasibility of adopting the RL-based model of substance addiction in natural agents in the analysis of wireheading in RL-based artificial agents. We presented an analytical extension to a TD-learning based model of addiction, and established sufficient parametric conditions on reward functions for the emergence of addictive behavior in AI agents. To verify this extension, we presented experimental results obtained from Q-learning agents learning to play the game of Snake, which is modified to include drug-induced surges in instantaneous rewards. The results demonstrate the promising potential of adopting the psychopathological models of mental disorders in the analysis of complex AI safety problems. 

%

%
%
%
%
\end{document}